\documentclass[10pt,twocolumn,letterpaper]{article}

\usepackage[pagenumbers]{wacv}  

\usepackage[accsupp]{axessibility}  
\usepackage{mathtools}
\usepackage{graphicx}
\usepackage{amsmath}
\usepackage{amssymb}
\usepackage{booktabs}

\usepackage{multirow}
\usepackage{pifont}
\newcommand{\cmark}{\ding{51}}%
\newcommand{\xmark}{\ding{55}}%
\usepackage[table]{xcolor}
\usepackage{wrapfig}
\usepackage[pagebackref,breaklinks,colorlinks]{hyperref}

\usepackage[capitalize]{cleveref}
\crefname{section}{Sec.}{Secs.}
\Crefname{section}{Section}{Sections}
\Crefname{table}{Table}{Tables}
\crefname{table}{Tab.}{Tabs.}

\def\wacvPaperID{1693} 
\def\confName{WACV}
\def\confYear{2024}

\begin{document}

\title{Link Prediction for Flow-Driven Spatial Networks}

\author{Bastian Wittmann\\
University of Zurich\\
{\tt\small bastian.wittmann@uzh.ch}
\and
Johannes C. Paetzold\\
Technical University of Munich\\
{\tt\small johannes.paetzold@tum.de}
\and
Chinmay Prabhakar\\
University of Zurich\\
{\tt\small chinmay.prabhakar@uzh.ch}
\and
Daniel Rueckert\\
Technical University of Munich\\
{\tt\small daniel.rueckert@tum.de}
\and
Bjoern Menze\\
University of Zurich\\
{\tt\small bjoern.menze@uzh.ch}
}
\maketitle

\begin{abstract}
Link prediction algorithms aim to infer the existence of connections (or links) between nodes in network-structured data and are typically applied to refine the connectivity among nodes. In this work, we focus on link prediction for flow-driven spatial networks, which are embedded in a Euclidean space and relate to physical exchange and transportation processes (e.g., blood flow in vessels or traffic flow in road networks). To this end, we propose the Graph Attentive Vectors (GAV) link prediction framework. GAV models simplified dynamics of physical flow in spatial networks via an attentive, neighborhood-aware message-passing paradigm, updating vector embeddings in a constrained manner. We evaluate GAV on eight flow-driven spatial networks given by whole-brain vessel graphs and road networks. GAV demonstrates superior performances across all datasets and metrics and outperformed the state-of-the-art on the ogbl-vessel benchmark at the time of submission by 12\% (98.38 vs. 87.98 AUC). All code is publicly available on GitHub.\footnote{\url{https://github.com/bwittmann/GAV}}
\end{abstract}

\vspace{-0.4em}
\section{Introduction and Motivation}\label{sec:intro}
\begin{figure}[t]
\centerline{\includegraphics[width=\linewidth]{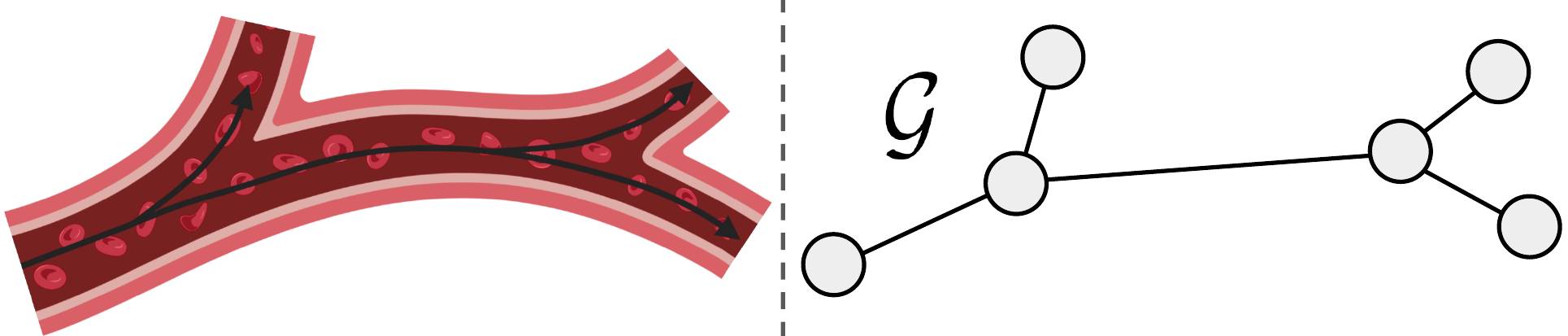}}
\caption{
Flow-driven spatial network $\mathcal{G}$ (right), representing vasculature (left). $\mathcal{G}$'s nodes are embedded in a Euclidean space and represent spatial positions specified by $x$-, $y$-, and $z$-coordinates.
}
\vspace{-0.6em}
\label{fig:intro}
\end{figure}

Networks (or graphs) can serve as efficient representations of real-world, ultra-complex systems and can be further classified into different categories. A prominent category is represented by undirected networks embedded in a Euclidean space constrained by geometry, called spatial networks~\cite{barthelemy2011spatial}. In this work, we are focusing on spatial networks, where a form of physical exchange or \emph{flow} can be used to describe characteristic functional properties of the underlying physical system. Examples include road networks, water bodies, and global exchange networks, but they can also be found in biology (\eg, vascular system, lymphatic system, and connectome). We will refer to such networks as \emph{flow-driven spatial networks} (see Fig.~\ref{fig:intro}).

Predominantly, network representations of physical systems originate from imaging methodologies, such as nanometer-scale microscopy in biology or regional to continental scale satellite remote sensing for road networks. The generation of compact network representations from these images is a multi-stage and imperfect process (segmentation, skeletonization, and subsequent graph pruning), which introduces noise and artifacts. However, for many relevant downstream tasks operating on flow-driven spatial networks, such as blood flow modeling~\cite{schmid2021severity} - a crucial component of investigating neurovascular brain disorders - or traffic forecasting~\cite{jiang2022graph}, a flawless network representation is of utmost importance. In this context, the task of link prediction presents itself as a meaningful measure to identify absent connections and reduce local noise arising from the erroneous graph generation process~\cite{paetzold2021whole, hu2020ogb}. Link prediction algorithms tailored to flow-driven spatial networks, however, remain heavily under-explored.

Therefore, we bring a simplistic yet general definition of the principle of physical flow, characterized by a direction and magnitude, to link prediction in graph representation learning. Our \textit{hypothesis} is that for flow-driven spatial networks, link prediction algorithms should heavily benefit from considering known functional properties, such as the aforementioned \emph{physical flow}, which are defined by the structural properties of the network (\eg, bifurcation angles~\cite{schneider2012tissue}). To this end, we propose the \emph{Graph Attentive Vectors} (GAV) link prediction framework. GAV operates on \emph{vector embeddings} representative of the network's structural properties and updates them in a constrained manner, imitating simplified dynamics of physical flow in spatial networks (\eg, blood flow in the vascular system or traffic flow in road networks).
We summarize our core contribution as follows:
\begin{enumerate}
\itemsep0em 
    \item We propose an attentive, neighborhood-aware message passing layer, called GAV layer, which updates vector embeddings, mimicking the (change in) direction and magnitude of physical flow in spatial networks.
     
    \item We introduce a readout module that aggregates vector embeddings in a physically plausible way and thus facilitates the interpretability of results.
     
    \item We prove the above-mentioned hypothesis by demonstrating state-of-the-art performance across all metrics in extensive experiments on eight flow-driven spatial networks, including the Open Graph
    Benchmark's~\cite{hu2020ogb} ogbl-vessel benchmark\footnote{\url{https://ogb.stanford.edu/docs/linkprop}} (98.38 vs. 87.98 AUC).  
\end{enumerate}

\section{Related Works}
This section discusses previous work on link prediction algorithms and message-passing layers.
Particular emphasis is placed on methods featured in our experiments.

\subsection{Link Prediction}
Link prediction algorithms are applied in various fields, such as social network analysis~\cite{murata2007link, liben2003link, daud2020applications}, bioinformatics~\cite{ lei2013novel, stanfield2017drug, kang2022lr}, recommender systems~\cite{ai2019link, talasu2017link, huang2005link, zhang2010solving}, and supply chain improvement~\cite{lu2020discovering}.
Broadly speaking, different link prediction algorithms try to estimate link existence between two nodes either via heuristic or learned methods. 

Heuristic algorithms employ predefined heuristics to encode the similarity between nodes. Some prominent candidates
are represented by common neighbors, resource allocation~\cite{zhou2009predicting}, preferential attachment~\cite{barabasi1999emergence}, Adamic-Adar~\cite{adamic2003friends}, Jaccard~\cite{jaccard1901etude}, Katz~\cite{katz1953new}, and average commute time~\cite{fouss2007random}. However, all heuristic link prediction algorithms suffer from the same underlying issue. They exploit predefined, simple heuristics, which can not be modified to account for different network types. \Eg, common neighbors has been developed for social networks and hence yields underwhelming results when applied to molecular graphs.

On the other hand, learned algorithms do not rely on predesigned heuristics but rather learn a more complex, data-driven heuristic utilizing neural networks. Thus, learned algorithms can easily adapt to different network types while typically outperforming their heuristic counterparts. SEAL~\cite{zhang2018link, zhang2021labeling} represents a prominent, learned link prediction framework, defining link prediction as a subgraph-level classification task by training a binary GNN-based classifier to map from subgraph patterns to link existence. To this end, SEAL first extracts a local subgraph around the link of interest, which is subsequently forwarded to DGCNN~\cite{zhang2018end} for classification. Moreover, SEAL's authors introduce an additional node labeling technique, known as labeling trick~\cite{zhang2021labeling}, to enhance the expressiveness of node features obtained from GNNs. SIEG~\cite{sieg} builds upon SEAL and introduces, inspired by Graphormer~\cite{ying2021transformers}, a pairwise structural attention module between two nodes of interest to capture local structural information more effectively. This results in state-of-the-art performances and allows SIEG to simultaneously overcome Graphormer's issue of exploding computational complexity when applied to ultra-large graphs. SUREL+~\cite{yin2023surel+} introduces the use of node sets to represent subgraphs. To complement the loss of structural information, SUREL+ provides set samplers, structure encoders, and set neural encoders. SUREL+ outperforms the baseline SEAL on various link prediction benchmarks. It should be mentioned that Cai \etal~\cite{cai2021line} investigate the use of line graphs for link prediction. Please note that none of the above-mentioned methods are tailored to flow-driven spatial networks.

\subsection{Message-Passing Layers}
GNNs utilize the concept of message-passing to encode semantically rich features within network-structured data. Over time, multiple variations of message-passing layers have been proposed~\cite{xu2018powerful, defferrard2016convolutional, fey2018splinecnn, he2020lightgcn}. For instance, GCN's message-passing layer~\cite{kipf2016semi} weighs each incoming message with a fixed coefficient, the node degree, before aggregation. In contrast, GAT's message-passing layer~\cite{brody2021attentive} learns aggregation weights dynamically based on attention scores. GraphSAGE's message-passing layer~\cite{hamilton2017inductive} does not directly aggregate central node features with incoming messages. Instead, it distinguishes these two kinds of features and learns two different transformations, one on the central node and another on incoming messages. EdgeConv~\cite{wang2019dynamic} aggregates the feature difference between the central node and its neighbors combined with the central node's features. Thus, EdgeConv draws parallels to aggregating spatial vectors if the nodes embed spatial positions.
However, our proposed GAV layer differs significantly from EdgeConv, as we explicitly constrain the update of vector embeddings to imitate the simplified dynamics of physical flow in flow-driven spatial networks. Importantly, only a few works tried to adapt the message-passing paradigm to spatial networks~\cite{zhang2021representation, danel2020spatial}.

\section{The Graph Attentive Vectors Framework}
\begin{figure*}[h]
\centerline{\includegraphics[width=\linewidth]{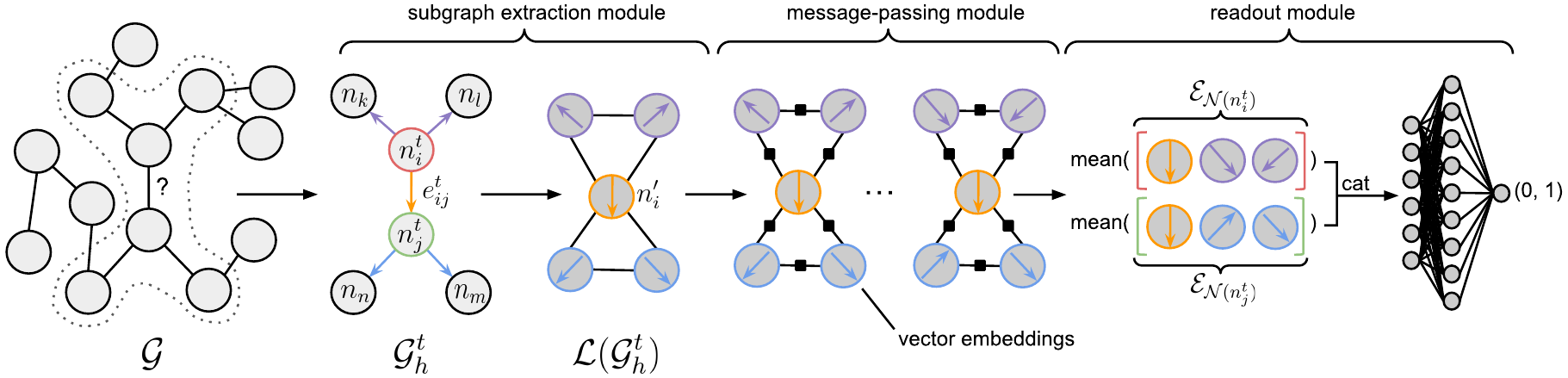}}
\caption{
Overview of the GAV link prediction framework. GAV is divided into three modules, namely the subgraph extraction module, the message-passing module, and the readout module.
First, an $h$-hop enclosing subgraph $\mathcal{G}^{t}_{h}$ is extracted around the target nodes $\{ n_{i}^{t}, n_{j}^{t} \}$ (red and green) affiliated to the target link $e^{t}_{ij}$ (orange) and subsequently transformed into a line graph representation $\mathcal{L}(\mathcal{G}^{t}_{h})$ to construct vector embeddings representative of the network's structural properties; second, we perform iterative message-passing among vector embeddings in $\mathcal{L}(\mathcal{G}^{t}_{h})$ via $k$ GAV layers, modeling simplified physical flow in spatial networks; and, third, a final subgraph-level readout module aggregates refined vector embeddings and predicts the probability of link existence with regard to the target link $e^{t}_{ij}$.
To provide a concise visualization, $h$ was set to 1. We would like to draw the reader's attention to color coding.
}
\label{fig:method_overview}
\end{figure*}

Our proposed Graph Attentive Vectors (GAV) link prediction framework is depicted in Fig.~\ref{fig:method_overview}. GAV represents a simple yet effective end-to-end trainable framework tailored to the task of link prediction for flow-driven spatial networks. It predicts the probability of link (or edge) existence between two nodes in a graph $\mathcal{G}$ based on a binary classifier $\mathcal{F}$, composed of a novel message-passing and readout module operating on vector embeddings (see Sections~\ref{ref:gav_layer} and~\ref{ref:readout}).
To this end, $\mathcal{F}$ should be able to differentiate between positive (real) and negative (sampled) links by assigning high probabilities of existence to plausible and low probabilities of existence to implausible links.
Following most competitive approaches~\cite{zhang2018link, zhang2021labeling, sieg}, we treat link prediction as a subgraph classification task. To determine the probability of existence of an individual target link between two target nodes, we, therefore, first extract an enclosing subgraph describing the target link's local neighborhood in a subgraph extraction module (see Section~\ref{ref:subnetw_extraction}). Subsequently, the subgraph is transformed into a line graph representation to construct vector embeddings and forwarded to $\mathcal{F}$, resulting in an iterative link prediction scheme predicting the existence of target links one at a time.
In the following, we elaborate extensively on GAV's individual components (see Fig.~\ref{fig:method_overview}).
For ease of reference, we provide a notation lookup table in the supplementary (see Supp.~\ref{sup:notation}).

\subsection{Subgraph Extraction Module}\label{ref:subnetw_extraction}
The undirected input graph $\mathcal{G}\coloneqq(\mathcal{V}, \mathcal{E})$ is defined by a set of nodes $\mathcal{V}$ and a set of corresponding edges $\mathcal{E}$. While a node $n_{i} \in \mathcal{V}$ contains a specific spatial position given by coordinates ($n_{i} \in \mathbb{R}^{d_\text{spatial}}$), an edge $e_{ij} \in \mathcal{E}$ indicates a connection between nodes $n_{i}$ and $n_{j}$.

Similar to most competitive approaches~\cite{zhang2018link, zhang2021labeling, sieg}, we extract as a first step an $h$-hop enclosing subgraph $\mathcal{G}^{t}_{h}$ around the nodes $\{ n_{i}^{t}, n_{j}^{t} \}$ affiliated to the target link $e^{t}_{ij}$ from the original graph representation $\mathcal{G}$ (please note that we refer to the target in our notations as $t$).
This results in an expressive and efficient representation of the target link's local neighborhood, including relevant structural patterns necessary to determine link existence. 
Further, the subgraph extraction significantly reduces space complexity, which is crucial for link prediction in ultra-large graphs (see Table~\ref{tab:datasets}).

Since GAV operates on \textit{vector embeddings}, we subsequently transform $\mathcal{G}^{t}_{h}$ into a line graph representation $\mathcal{L}(\mathcal{G}^{t}_{h})\coloneqq(\mathcal{V'}, \mathcal{E'})$. In $\mathcal{L}(\mathcal{G}^{t}_{h})$, each node $n'_{i} \in \mathcal{V'}$ represents an edge $e_{ij} \in \mathcal{E}$ via an individual vector embedding, while an edge $e'_{ij} \in \mathcal{E'}$ indicates adjacency between two edges in $\mathcal{G}^{t}_{h}$ iff they are incident. To create vector embeddings, we encode edges as vectors between the involved nodes (\eg, $n'_{i} = n_{j} - n_{i}$). Therefore, $\mathcal{L}(\mathcal{G}^{t}_{h})$'s node embeddings are defined as vectors (see Fig.~\ref{fig:method_overview}) of unique length and direction representative of edges in $\mathcal{G}^{t}_{h}$ (see Supp.~\ref{sup:imp_det} for implementation details). The line graph $\mathcal{L}(\mathcal{G}^{t}_{h})$ formed around the target link $e^{t}_{ij}$ is finally forwarded to the message-passing module.

\subsection{Message-Passing Module and GAV Layer}\label{ref:gav_layer}
\begin{figure}[h]
\centerline{\includegraphics[width=0.95\linewidth]{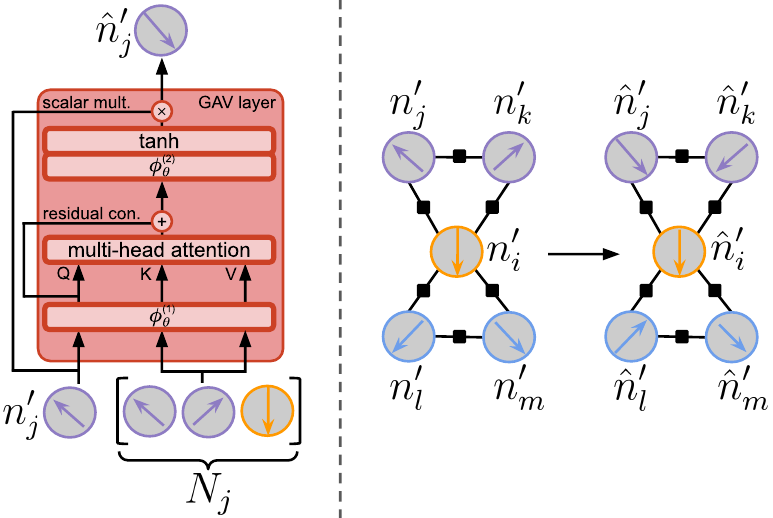}}
\caption{
Visualization of a single GAV layer updating the vector embedding of node $n'_{j}$. We forward the vector embedding of node $n'_{j}$ together with $N_{j} \in \mathbb{R}^{| \mathcal{N}(n'_j) \cup {n'_j}| \times d_{\text{spatial}}}$, which represents the set of $n'_{j}$ and its neighbors $n'_{k}$ and $n'_{i}$, to the GAV layer.
In this specific example, the vector embedding has been flipped (see $\hat{n}'_{j}$), which can be interpreted as a change in direction of physical flow.
}
\label{fig:mpn}
\end{figure}

To adjust link prediction to flow-driven spatial networks, we propose a novel message-passing layer, termed GAV layer. We perform $k$ iterations of message-passing among the nodes of the line graph $\mathcal{L}(\mathcal{G}^{t}_{h})$, obtained from our subgraph extraction module, in our message-passing module. The GAV layer's message-passing relies on a straightforward intuition inspired by principles of physical flow. To be precise, we treat nodes in $\mathcal{L}(\mathcal{G}^{t}_{h})$ as vector embeddings and update them in a constrained manner, imitating simplified dynamics of physical flow in spatial networks. The detailed structure of a single GAV layer is visualized in Fig.~\ref{fig:mpn}.

In order to update an individual vector embedded in a node $n'_{i} \in \mathbb{R}^{d_{\text{spatial}}}$, we first project it together with a matrix $N_{i} \in \mathbb{R}^{| \mathcal{N}(n'_{i}) \cup n'_{i}| \times d_{\text{spatial}}}$, consisting of directly neighboring nodes and the node itself, 
into a higher dimensional space $d_{\text{message}}$; the projection is through a learnable function $\phi^{(1)}_{\theta}:\mathbb{R}^{d_{\text{spatial}}} \rightarrow \mathbb{R}^{d_{\text{message}}}$. Subsequently, $\phi^{(1)}_{\theta}(n'_{i})$ and $\phi^{(1)}_{\theta}(N_{i})$ are forwarded to a multi-head attention operation, where $\phi^{(1)}_{\theta}(n'_{i})$ represents a single query, while $\phi^{(1)}_{\theta}(N_{i})$ represents the key and value sequence.
\begin{equation}
\tilde{n}'_{i} = \text{MultiHeadAttn}(\phi^{(1)}_{\theta}(n'_{i}),\ \phi^{(1)}_{\theta}(N_{i}),\ \phi^{(1)}_{\theta}(N_{i}))
\end{equation}
This results in an intermediate node representation $\tilde{n}'_{i} \in \mathbb{R}^{d_{\text{message}}}$, which incorporates not only information of the node itself but also its local structural neighborhood via the concept of attention. In the next step, we apply a residual connection for increased gradient flow and forward the result to the learnable function $\phi^{(2)}_{\theta}: \mathbb{R}^{d_{\text{message}}} \rightarrow \mathbb{R}$, followed by a tanh non-linearity.
\begin{equation}
s_{i} = \text{tanh}(\phi^{(2)}_{\theta}(\tilde{n}'_{i} + \phi^{(1)}_{\theta}(n'_{i})))
\end{equation}
Finally, the scalar value $s_{i} \in (-1, 1)$ is utilized to update the original node representation $n'_{i}$ via scalar multiplication.
\begin{equation}
\hat{n}'_{i} =  s_{i} \cdot n'_{i}
\end{equation}
Hence, after one layer of message-passing, the updated, refined node representation is given by $\hat{n}'_{i} \in \mathbb{R}^{d_{\text{spatial}}}$. Importantly, $\hat{n}'_{i}$ preserves relevant structural properties of the original node representation $n'_{i}$.
This is because scalar multiplication with $s_{i} \in (-1, 1)$ restricts the modification of vector embeddings given by nodes in $\mathcal{L}(\mathcal{G}^{t}_{h})$.
In essence, our message-passing paradigm can be geometrically interpreted as a scaling combined with a potential flipping operation of vectors. These constraints imposed by our message-passing align with our principal idea of modeling simplified dynamics of physical flow in spatial networks via potential, constrained changes in the direction and magnitude of vector embeddings, preserving the network's structural properties.

\begin{table*}[h] 
\centering
\scriptsize
\caption{Properties of the raw datasets. Each dataset consists of exactly one ultra-large graph.}
\vspace{-0.5em}
\label{tab:datasets}
\begin{tabular}{l |l l c l c c l} 
\toprule
 Dataset Name & $\#$ Nodes & $\#$ Edges & Node Degree & Node Features & $d_\text{spatial}$ & Edge Features & Description\\
\midrule
ogbl-vessel~\cite{hu2020ogb} & 3,538,495 & 5,345,897 & 3.02 & $x$-, $y$-, $z$-coordinates & 3 & $-$ & BALB/c mouse $\text{strain}^{1}$\\
c57-tc-vessel~\cite{paetzold2021whole} & 3,820,133 & 5,614,677 & 2.94 & $x$-, $y$-, $z$-coordinates & 3 & $-$ & C57BL/6 mouse $\text{strain}^{1}$\\
cd1-tc-vessel~\cite{paetzold2021whole} & 3,645,963  & 5,791,309 & 3.18 & $x$-, $y$-, $z$-coordinates & 3 & $-$ & CD-1 mouse $\text{strain}^{1}$\\
c57-cc-vessel~\cite{walchli2021hierarchical} & 6,650,580 & 9,054,100 & 2.72 & $x$-, $y$-, $z$-coordinates &3 &  $-$ & C57BL/6 mouse $\text{strain}^{2}$\\
\midrule
belgium-road~\cite{bader201110th} & 1,441,295 & 1,549,970 & 2.15 & $x$-, $y$-coordinates & 2 & $-$ & Belgium\\
italy-road~\cite{bader201110th} & 6,686,493 & 7,013,978 & 2.10 & $x$-, $y$-coordinates & 2 & $-$ & Italy\\
netherlands-road~\cite{bader201110th} & 2,216,688 & 2,441,238 & 2.20& $x$-, $y$-coordinates & 2 & $-$ & Netherlands\\
luxembourg-road~\cite{bader201110th} & 114,599 & 119,666 & 2.09 & $x$-, $y$-coordinates & 2 & $-$ & Luxembourg\\
\bottomrule
\multicolumn{7}{l}{$^{1}$ tissue clearing (tc) and light-sheet microscopy imaging  $\qquad ^{2}$ corrosion casting (cc) and SR\textmu$\text{CT}$ imaging}
\end{tabular}
\end{table*}

\subsection{Labeling Trick}
\begin{figure}[h]
\centerline{\includegraphics[width=0.45\linewidth]{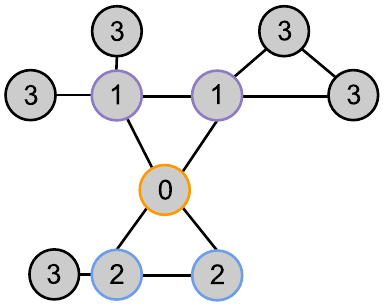}}
\caption{
Labels generated by the labeling trick for an exemplary line graph ($h$ set to two).
Our interpretation of the labeling trick assigns the label 0 to the vector embedding representing the target link (orange), the labels 1 and 2 to vector embeddings representing edges connected to the target nodes $n^t_{i}$ and $n^t_{j}$ (purple and blue), and the label 3 to remaining vector embeddings.
}
\label{fig:labeling_trick}
\end{figure}

Following common practices~\cite{zhang2021labeling, sieg, cai2021line}, we apply a labeling trick to enable the message-passing module to distinguish between the relevance of different vector embeddings and, hence, learn an expressive structural representation of the target link's local neighborhood.
The labeling trick ensures that vector embeddings created from the target link $e_{ij}^{t}$ and edges connected to the target nodes $\{ n_{i}^{t}, n_{j}^{t} \}$ are identifiable by a distinct label.
Labels generated by our interpretation of the labeling trick are shown in Fig.~\ref{fig:labeling_trick}.

\noindent
The additional labels generated by the labeling trick are concatenated to $\mathcal{L}(\mathcal{G}^{t}_{h})$'s vector embeddings in the message-passing module (before message-passing).

\subsection{Readout Module}\label{ref:readout}
Our proposed readout module consists of a custom node aggregation operation followed by a learnable function $\phi^{(3)}_{\theta}: \mathbb{R}^{2 \cdot d_{\text{spatial}}} \rightarrow \mathbb{R}$ and processes refined vector embeddings obtained from the message-passing module. 
While the node aggregation operation aims to distill pertinent information from the refined graph representation in a fashion invariant to node ordering and quantity, $\phi^{(3)}_{\theta}$ predicts the probability of link existence with regard to the target link.

Equation~\ref{eq:agg} summarizes the readout module's functionality, where $\mathcal{E}_{\mathcal{N}(n^t_{i})}$ defines the set of refined vector embeddings originally created from edges adjacent to $n^t_{i}$ (see Fig.~\ref{fig:method_overview}), $\mathcal{E}_{\mathcal{N}(n^t_{j})}$ the set of refined vector embeddings originally created from edges adjacent to $n^t_{j}$ (see Fig.~\ref{fig:method_overview}), $\mathbin\Vert$ the concatenation operation, and $\hat{y}^t_{ij}$ the predicted probability of target link existence.
\begin{equation} \label{eq:agg}
\hat{y}^t_{ij} =  \phi^{(3)}_{\theta}(\text{mean}(\mathcal{E}_{\mathcal{N}(n^t_{i})}) \mathbin\Vert \text{mean}(\mathcal{E}_{\mathcal{N}(n^t_{j})}))
\end{equation}
Thus, our node aggregation operation constructs mean vector embeddings (see \text{mean}($\mathcal{E}_{\mathcal{N}(n^t_{i})})$ and $\text{mean}(\mathcal{E}_{\mathcal{N}(n^t_{j})})$) representative of our simplified definition of flow in the two target nodes and intends to exploit their relationship to predict whether target nodes should be connected or not.

\subsection{Loss Function}
Since our approach represents a binary classifier $\mathcal{F}$ determining the probability of link existence with regard to the target link, we optimize a binary cross-entropy loss function during training. Here, $y^t_{ij} \in \{0, 1\}$ indicates whether the target links are negative (sampled) or positive (real).
\begin{equation}
\mathcal{L_{\text{BCE}}} =  \frac{-1}{| \mathcal{E} |} \sum_{ij \in \mathcal{E}} y^t_{ij} \cdot \text{log}(\hat{y}^t_{ij}) + (1 - {y}^t_{ij}) \cdot \text{log}(1 - \hat{y}^t_{ij})
\end{equation}
We would like to highlight that our approach is trained in an entirely end-to-end manner, solely based on the information of link existence. Therefore, intermediate representations, such as the refined vector embeddings, are determined completely data-driven.

\section{Experiments and Results}
\begin{table*}[t]
\centering
\scriptsize
\caption{Quantitative results.
GAV outperforms the previous state-of-the-art across all metrics and datasets. We report mean and standard deviation values on the ogbl-vessel benchmark based on ten different random seeds (0 - 9). To compare GAV to existing baseline algorithms, we report quantitative results based on the area under the receiver operating characteristic curve (AUC). We additionally utilize the evaluation metric Hits@$k$ as an additional, stricter performance measure. More details on evaluation metrics can be found in the supplementary (see Supp.~\ref{sup:metrics}). Please note that the ogbl-vessel benchmark's official evaluation metric is AUC. Therefore, Hits@$k$ values of participating algorithms are not available. We highlight the respective three best quantitative values in shades of teal.}
\label{tab:quantitative_results}
\vspace{-0.5em}
\begin{tabular}{p{50pt}|l|l|c|c|c|c} 
\toprule
Dataset & Model &$\#\ \text{Params}\downarrow$ &$\text{AUC}\uparrow$ (\%)& $\text{Hits@100}\uparrow$  (\%)& $\text{Hits@50}\uparrow$  (\%)& $\text{Hits@20}\uparrow  (\%)$\\ 
\midrule
\multirow{13}{*}{ogbl-vessel}
& GCN~\cite{kipf2016semi} & 396,289 & 43.53 ± 9.61 & - & - & - \\
& MLP & 1,037,577 & 47.94 ± 1.33 & - & - & - \\
& Adamic-Adar~\cite{adamic2003friends} &  \cellcolor{teal!40}0 & 48.49 ± 0.00 & - & - & - \\
& GraphSAGE~\cite{hamilton2017inductive} & 396,289 & 49.89 ± 6.78 & - & - & - \\
& SAGE+JKNet~\cite{xu2018representation} & \cellcolor{teal!20}273 & 50.01 ± 0.07 & - & - & - \\
& SGC~\cite{wu2019simplifying} & \cellcolor{teal!10}897 & 50.09 ± 0.11 & - & - & - \\
& LRGA~\cite{puny2020global} & 26,577 & 54.15 ± 4.37 & - & - & - \\
& SEAL~\cite{zhang2021labeling} & 172,610 & 80.50 ± 0.21 & - & - & - \\
& S3GRL ($\text{PoS}^{\text{+}}$)~\cite{louis2023simplifying} & 2,382,849 & 80.56 ± 0.06 & - & - & - \\
& SUREL+~\cite{yin2023surel+} & 56,353 & 84.96 ± 0.68 & - & - & - \\
& SIEG~\cite{sieg} & 752,716 & \cellcolor{teal!10}87.98 ± 1.00 & - & - & - \\
\cmidrule{2-7}
& SEAL+EdgeConv (ours) & 49,346 & \cellcolor{teal!20}97.53 ± 0.32 & \cellcolor{teal!20}16.09 ± 10.48 & \cellcolor{teal!20}9.37 ± 6.18 & \cellcolor{teal!20}4.99 ± 4.24\\
& GAV (ours) & 8,194 & \cellcolor{teal!40}98.38 ± 0.02 & \cellcolor{teal!40}34.77 ± 0.94 & \cellcolor{teal!40}28.02 ± 1.58 & \cellcolor{teal!40}19.71 ± 2.31\\
\midrule
\multirow{3}{*}{c57-tc-vessel}
& SEAL~\cite{zhang2021labeling} & \cellcolor{teal!10}43,010 & \cellcolor{teal!10}78.21 & \cellcolor{teal!10}0.12 & \cellcolor{teal!10}0.06 & \cellcolor{teal!10}0.01\\
& SEAL+EdgeConv (ours) & \cellcolor{teal!20}49,346 & \cellcolor{teal!20}97.23 & \cellcolor{teal!20}16.71 & \cellcolor{teal!20}10.39 & \cellcolor{teal!20}5.01\\
& GAV (ours) & \cellcolor{teal!40}8,194 & \cellcolor{teal!40}98.24 & \cellcolor{teal!40}33.26 & \cellcolor{teal!40}26.89 & \cellcolor{teal!40}21.32\\
\midrule
\multirow{3}{*}{cd1-tc-vessel}
& SEAL~\cite{zhang2021labeling} & \cellcolor{teal!10}43,010 & \cellcolor{teal!10}83.60 & \cellcolor{teal!10}0.27 & \cellcolor{teal!10}0.16 & \cellcolor{teal!10}0.06\\
& SEAL+EdgeConv (ours) & \cellcolor{teal!20}49,346 & \cellcolor{teal!20}97.91 & \cellcolor{teal!20}17.05 & \cellcolor{teal!20}11.57 & \cellcolor{teal!20}2.98\\
& GAV (ours) & \cellcolor{teal!40}8,194 & \cellcolor{teal!40}98.72 & \cellcolor{teal!40}35.82 & \cellcolor{teal!40}27.25 & \cellcolor{teal!40}17.23\\
\midrule
\multirow{3}{*}{c57-cc-vessel}
& SEAL~\cite{zhang2021labeling} & \cellcolor{teal!10}43,010 & \cellcolor{teal!10}83.75 & \cellcolor{teal!10}0.65 & \cellcolor{teal!10}0.44 & \cellcolor{teal!10}0.24\\
& SEAL+EdgeConv (ours) & \cellcolor{teal!20}49,346 & \cellcolor{teal!20}97.49 & \cellcolor{teal!20}7.21 & \cellcolor{teal!20}3.35 & \cellcolor{teal!20}1.06\\
& GAV (ours) & \cellcolor{teal!40}8,194 & \cellcolor{teal!40}97.99 & \cellcolor{teal!40}18.90 & \cellcolor{teal!40}14.58 & \cellcolor{teal!40}9.04\\
\midrule
\multirow{3}{*}{belgium-road}
& SEAL~\cite{zhang2021labeling} & \cellcolor{teal!10}43,010 & \cellcolor{teal!10}86.73 & \cellcolor{teal!20}1.25 & \cellcolor{teal!20}0.68 & \cellcolor{teal!10}0.30\\
& SEAL+EdgeConv (ours) & \cellcolor{teal!20}49,346 & \cellcolor{teal!20}96.98 & \cellcolor{teal!10}0.55 & \cellcolor{teal!10}0.55 & \cellcolor{teal!20}0.51\\
& GAV (ours) & \cellcolor{teal!40}8,194 & \cellcolor{teal!40}99.29 & \cellcolor{teal!40}47.44 & \cellcolor{teal!40}38.60 & \cellcolor{teal!40}22.11\\
\midrule
\multirow{3}{*}{italy-road}
& SEAL~\cite{zhang2021labeling} & \cellcolor{teal!10}43,010 &  \cellcolor{teal!10}90.07 &  \cellcolor{teal!20}0.32 &  \cellcolor{teal!10}0.16 & \cellcolor{teal!20}0.08\\
& SEAL+EdgeConv (ours) & \cellcolor{teal!20}49,346 & \cellcolor{teal!20}90.24 & \cellcolor{teal!10}0.26 & \cellcolor{teal!20}0.17 &  \cellcolor{teal!10}0.07\\
& GAV (ours) & \cellcolor{teal!40}8,194 & \cellcolor{teal!40}99.41 & \cellcolor{teal!40}28.49 & \cellcolor{teal!40}20.08 & \cellcolor{teal!40}11.99\\
\midrule
\multirow{3}{*}{netherlands-road}
& SEAL~\cite{zhang2021labeling} & \cellcolor{teal!10}43,010 & \cellcolor{teal!10}84.19 & \cellcolor{teal!10}0.00 & \cellcolor{teal!10}0.00 & \cellcolor{teal!10}0.00\\
& SEAL+EdgeConv (ours) & \cellcolor{teal!20}49,346 & \cellcolor{teal!20}96.06 & \cellcolor{teal!20}3.91 & \cellcolor{teal!20}2.20 & \cellcolor{teal!20}1.01\\
& GAV (ours) & \cellcolor{teal!40}8,194 & \cellcolor{teal!40}99.44 & \cellcolor{teal!40}37.55 & \cellcolor{teal!40}26.97 & \cellcolor{teal!40}10.77\\
\midrule
\multirow{3}{*}{luxembourg-road}
& SEAL~\cite{zhang2021labeling} & \cellcolor{teal!10}43,010 & \cellcolor{teal!10}89.79 & \cellcolor{teal!10}11.39 & \cellcolor{teal!10}6.15 & \cellcolor{teal!10}3.12\\
& SEAL+EdgeConv (ours) & \cellcolor{teal!20}49,346 & \cellcolor{teal!20}97.53 & \cellcolor{teal!20}59.79 & \cellcolor{teal!20}39.15 & \cellcolor{teal!20}19.42\\
& GAV (ours) & \cellcolor{teal!40}8,194 & \cellcolor{teal!40}99.31 & \cellcolor{teal!40}85.88 & \cellcolor{teal!40}76.84 & \cellcolor{teal!40}61.95\\
\bottomrule
\end{tabular}
\end{table*}

In this section, we demonstrate the performance of our proposed GAV framework on the ogbl-vessel benchmark~\cite{hu2020ogb} and on additional datasets sourced from publicly available flow-driven spatial networks.
We first elaborate on baseline algorithms and the experimental setup, followed by a detailed description of datasets utilized in our experiments. Finally, we report quantitative results, investigate our design choices by conducting extensive ablation studies, and discuss GAV's interpretability in detail.
Additional experiments on non-flow-based benchmarks and roto-translational invariance can be found in the supplementary.

\subsection{Baselines and Experimental Setup}\label{ref:baselines}
To evaluate GAV properly, we not only compare GAV to algorithms submitted to the ogbl-vessel benchmark but also propose a novel \emph{secondary baseline} (SEAL+EdgeConv) combining the SEAL framework~\cite{zhang2018link, zhang2021labeling} with the EdgeConv message-passing layer~\cite{wang2019dynamic}, following recent trends in graph-based object detection from point clouds~\cite{chai2021point, yang2022graph, wang2021object, yin2020lidar}. Extensive experiments with different baseline algorithms revealed that this provides us with an improved, highly competitive secondary baseline for link prediction in spatial networks. We, therefore, compare GAV on the additionally sourced datasets to SEAL's original version (with tuned hyperparameters), which has shown to deliver results on par with or superior to the state-of-the-art on multiple link prediction benchmarks, and to our introduced secondary baseline SEAL+EdgeConv. Details on the configuration of our secondary baseline can be found in the supplementary (see Supp.~\ref{sup:sec_base}).

GAV was trained using the Adam optimizer~\cite{kingma2014adam} with a learning rate of 0.001 and a batch size of 32 on a single Tesla V100 GPU (32 GB) until convergence.
An ablation study on the number of hops $h$ in the subgraph extraction module and the number of message-passing iterations $k$ indicates that setting both to one is sufficient (see Table \ref{tab:ablations_kh}).
In the GAV layer, the number of heads of the multi-head attention operation is set to 4, while $\phi^{(1)}_{\theta}$ represents a single linear layer with an output dimension of $d_{\text{message}} = 32$, and $\phi^{(2)}_{\theta}$ is given by a two-layer MLP with a hidden dimension of 64.
The GAV layer makes use of leaky ReLU non-linearities~\cite{maas2013rectifier} to increase gradient flow and simplify weight initialization. The readout module's learnable function $\phi^{(3)}_{\theta}$ is represented by a two-layer MLP with a hidden dimension of 128.
All hyperparameters were tuned on the validation set of the ogbl-vessel benchmark.

\subsection{Datasets}
We experiment with multiple 2D and 3D flow-driven spatial networks to demonstrate the generalizability of our approach (see Table~\ref{tab:datasets}). In total, we conduct experiments on eight networks, including the ogbl-vessel benchmark and seven additionally sourced datasets, representing whole-brain vessel graphs of different mouse strains and road networks of various European countries. In this context, link prediction can be interpreted as predicting the probability of the existence of blood vessels and road segments. Whole-brain vessel graphs and road networks are graphically visualized in the supplementary (see Supp.~\ref{sup:data}).

\paragraph{Whole-Brain Vessel Graphs}
Blood vessels represent fascinating structures forming complex networks that transport oxygen and nutrients throughout the human body. The vascular system is, therefore, intuitively represented as a flow-driven spatial network, where branching points of vessels typically represent nodes embedding $x$-, $y$-, and $z$-coordinates, while edges are defined as blood vessels running between branching points~\cite{paetzold2021whole}. We report results on the Open Graph Benchmark's ogbl-vessel benchmark~\cite{hu2020ogb}, which measures the performance of different link prediction algorithms with regard to whole-brain vessel graphs aiming to remove artifacts introduced by the multi-stage graph generation process. The ogbl-vessel benchmark consists of millions of nodes and edges (see Table~\ref{tab:datasets}) and describes the murine brain vasculature all the way down to the microcapillary level.
However, we not only experiment with the ogbl-vessel benchmark but also source three additional whole-brain vessel graphs of different mouse strains acquired via different imaging methodologies~\cite{todorov2020machine, walchli2021hierarchical} (see Table~\ref{tab:datasets}, footnote).

\paragraph{Road Networks}
Further, we report results on diverse road networks representative of four European countries for a thorough evaluation of our proposed GAV framework's generalizability. To this end, we adopt publicly available road networks introduced in the DIMACS graph partitioning and clustering challenge~\cite{bader201110th}. These road networks correspond to the largest connected components of OpenStreetMap's~\cite{OpenStreetMap} road networks and are vastly different in size (\eg, luxembourg-road constitutes roughly 100,000 edges, whereas italy-road has more than 7,000,000). 
In road networks, intersections and locations with stronger curvature represent nodes in the form of $x$- and $y$-coordinates, while connecting roads represent edges.

\paragraph{Preprocessing}
Link prediction datasets require positive (label 1) and negative links (label 0). Positive links correspond to existent edges in our datasets, whereas negative links represent artificially created, non-existent edges. As link prediction algorithms are commonly employed to improve the graph representation through the identification of absent connections and the reduction of local noise arising from graph generation, negative links should appear as authentic as possible.
In light of the absence of negative links in our sourced datasets, we prepare our sourced datasets in a manner that aligns with the ogbl-vessel benchmark. Therefore, we sample negative links using a spatial sampling strategy. To be precise, we randomly connect nodes in close proximity, taking a maximum distance threshold of $\delta = \overline{e_{ij}} + 2 \sigma $ into account. Here, $\overline{e_{ij}}$ denotes the average edge length estimated over the entire graph $\mathcal{G}$ and $\sigma$ the standard deviation. 
The number of negative, sampled links corresponds to the number of positive, real links across all datasets. 
We finally split positive and negative links into training, validation, and test sets (split 80\%/10\%/10\%).

\subsection{Quantitative Results}
GAV demonstrates excellent, superior performances on the task of link prediction across all metrics and datasets, as can be observed in Table~\ref{tab:quantitative_results}.
We outperform the current state-of-the-art algorithm SIEG on the ogbl-vessel benchmark by \textbf{12\% (98.38 vs. 87.98 AUC)} while requiring a significantly smaller amount of parameters (8,194 vs. 752,716). However, GAV not only drastically outperforms all algorithms submitted to the ogbl-vessel benchmark but also our introduced strong, secondary baseline, combining the SEAL framework with EdgeConv (SEAL+EdgeConv). The excellent performance and superiority of GAV is even more pronounced when considering the strict evaluation metric of Hits@$k$.
Quantitative results reported in Table~\ref{tab:quantitative_results} additionally indicate the strong performance of our introduced secondary baseline (see Section~\ref{ref:baselines}).

It is of note that the luxembourg-road dataset's test set contains only 12,000 negative links. We, therefore, compare predictions of its positive links to 12,000 rather than 100,000 negative links (see supplementary, Supp.~\ref{sup:metrics}). This explains the comparatively strong Hits@$k$ performances on the luxembourg-road dataset.

\subsection{Ablation Studies}
To further validate GAV, we conduct detailed ablation studies on the validation set of the ogbl-vessel benchmark. Additional ablation studies on the GAV layer and its design choices can be found in the supplementary (Supp.~\ref{sup:gav_abl}).
Table~\ref{tab:ablations_des} investigates the importance of the readout module, the message-passing module, and the labeling trick.
\begin{table}[h]
\centering
\scriptsize
\caption{Ablations on main design choices.}
\vspace{-0.5em}
\label{tab:ablations_des}
\begin{tabular}{c c c|c c} 
\toprule
Readout Module & Message-Passing & Labeling Trick & $\text{AUC}\uparrow$ &$\Delta$ \\
\midrule
\cmark & \cmark & \cmark &  \cellcolor{teal!40}98.39 & $-$\\
\xmark & \cmark & \cmark & \cellcolor{teal!20}98.28 & -0.11\\
\cmark & \xmark & \cmark & 80.56 & -17.83\\
\cmark & \cmark & \xmark & \cellcolor{teal!10}96.00 & -2.39\\
\bottomrule
\end{tabular}
\end{table}
First, we exchange our readout module with a SortPooling layer followed by two convolutional layers and an MLP, resembling SEAL's readout operation. We note that our readout module is more applicable to flow-driven spatial networks, as it leads to a modest AUC increase of 0.11. Second, we completely deactivate the message-passing module by forwarding $\mathcal{L}(\mathcal{G}^{t}_{h})$ directly to the readout module. We observe a drastic AUC decrease of 17.83, indicating the importance of modifying the vector embeddings via our proposed GAV layer.
Finally, we evaluate the impact of the labeling trick. Excluding the additional labels generated by the labeling trick results in an AUC decrease of 2.39, proving the significance of link identification via additional, distinct labels.

In a second ablation study, we experiment with different message-passing layers (see Supp.~\ref{sup:update}), including EdgeConv, in our message-passing module. We report our findings in Table~\ref{tab:ablations_operators}.
\begin{table}[h]
\centering
\scriptsize
\caption{Ablations with different message-passing layers.}
\label{tab:ablations_operators}
\vspace{-0.5em}
\begin{tabular}{c |c c c c} 
\toprule
Message-Passing Layer & $\text{AUC}\uparrow$ & $\text{Hits@100}\uparrow$& $\text{Hits@50}\uparrow$& $\text{Hits@20}\uparrow$ \\
\midrule
GAV layer (ours) &\cellcolor{teal!40}98.39 & \cellcolor{teal!40}34.46 & \cellcolor{teal!40}26.30 & \cellcolor{teal!40}19.81\\
EdgeConv~\cite{wang2019dynamic} &\cellcolor{teal!20}97.43 & \cellcolor{teal!20}17.30 & \cellcolor{teal!20}5.97 & \cellcolor{teal!10}0.78\\
GAT layer~\cite{brody2021attentive} &\cellcolor{teal!10}96.44 & \cellcolor{teal!10}4.58 & \cellcolor{teal!10}2.55 & \cellcolor{teal!20}1.59\\
SAGE layer~\cite{hamilton2017inductive} &93.53 & 0.77 & 0.11 & 0.03\\
GCN layer~\cite{kipf2016semi} &89.31 & 0.39 & 0.22 & 0.16\\
\bottomrule
\end{tabular}
\end{table}
Our proposed GAV layer outperforms the other message-passing layers across all metrics by a considerable amount, proving our rationale.

Lastly, we vary the number of hops $h$ used to generate $\mathcal{G}^t_h$ and the number of message-passing iterations $k$ (see Table~\ref{tab:ablations_kh}).
\begin{table}[h]
\centering
\scriptsize
\caption{Ablations on $k$ and $h$.}
\label{tab:ablations_kh}
\vspace{-0.5em}
\begin{tabular}{c |c c c c} 
\toprule
$k$ \& $h$ & $\text{AUC}\uparrow$ &$\text{Hits@100}\uparrow$& $\text{Hits@50}\uparrow$& $\text{Hits@20}\uparrow$\\
\midrule
1 & \cellcolor{teal!40}98.39 & \cellcolor{teal!40}34.46 & \cellcolor{teal!20}26.30 & \cellcolor{teal!20}19.81\\
2 & \cellcolor{teal!40}98.39 & \cellcolor{teal!10}34.00 & \cellcolor{teal!40}26.93 & \cellcolor{teal!40}21.21\\
3 & \cellcolor{teal!40}98.39 & \cellcolor{teal!20}34.25 & \cellcolor{teal!10}25.99 & \cellcolor{teal!10}17.84\\
\bottomrule
\end{tabular}
\end{table}
We observe that simultaneously increasing $k$ and $h$ results in no discernible differences in performance. This finding is in line with the $\gamma$-decaying theory~\cite{zhang2018link}, proving the approximability of high-order heuristics from locally restricted subgraphs.

\begin{figure*}[h]
\centerline{\includegraphics[width=\linewidth]{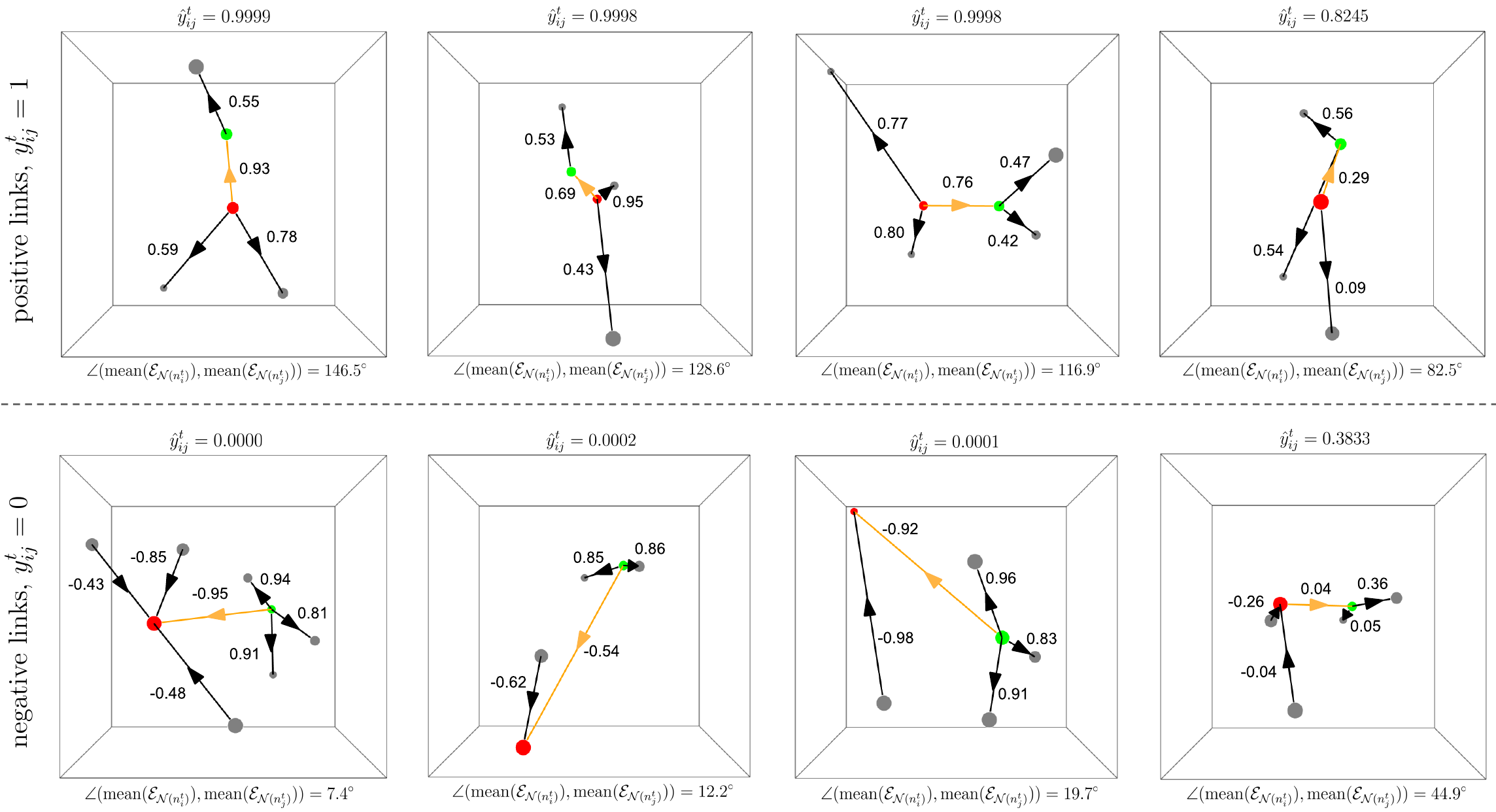}}
\vspace{-0.4em}
\caption{
Visualization of the effect of our GAV layer on vector embeddings (ogbl-vessel). We visualize subgraph representations $\mathcal{G}^t_h$ ($h$ set to one) of four positive, plausible (first row) and four negative, implausible target links (second row) together with the GAV layer's predicted scalar values $s_{i} \in (-1, 1)$. The scalar values $s_{i}$ used to update vector embeddings in $\mathcal{L}(\mathcal{G}^t_h)$ have been projected to $\mathcal{G}^t_h$ to provide an interpretable visualization. The directionality of edges already incorporates potential shifts in direction enforced by our GAV layer. Please note that following the color coding scheme of Fig.~\ref{fig:method_overview}, the target link $e^{t}_{ij}$ is depicted in orange, whereas the two target nodes $n_{i}^{t}$ and $n_{j}^{t}$ are displayed in red and green. We additionally report the angle $\angle$ between vector embeddings aggregated around the two target nodes (see Section~\ref{ref:readout}) and the predicted probability of link existence $\hat{y}^t_{ij}$. The last column contains more challenging cases.
}
\label{fig:inter}
\end{figure*}

\subsection{Interpretability and Analysis of Results}\label{ref:inter}
In Fig.~\ref{fig:inter}, we visualize the behavior of our proposed GAV layer, attempting to facilitate interpretability and provide a clearer insight on how it creates superior representations for link prediction in flow-driven spatial networks. An analysis based on a toy example and additional visualizations similar to Fig.~\ref{fig:inter} are provided in the supplementary (Supp~\ref{sup:toy} \&~\ref{sup:inter}).

Scalar multiplication via $s_{i}$ enables GAV to flip individual vector embeddings (mimicking \textit{change in direction of flow}). We observe that this results in physically implausible, learned sink/source flow between the two target nodes (red and green) of implausible formations (see Fig.~\ref{fig:inter}, second row; Fig.~\ref{fig:neg_preds}, supplementary) with a consistency of 94.26\%, which stands in drastic contrast to the behavior of physical flow in spatial networks. The learned sink/source flow between target nodes of negative links, in turn, leads to drastically different representations for implausible and plausible formations (see Fig.~\ref{fig:inter}, first row) that can be effortlessly classified in our readout module. The difference in representation is also reflected in the angle $\angle$ between vector embeddings aggregated around the two target nodes represented by $\text{mean}(\mathcal{E}_{\mathcal{N}(n^t_{i})})$ and $\text{mean}(\mathcal{E}_{\mathcal{N}(n^t_{j})})$, constructed in our readout module (see Section~\ref{ref:readout}). We find that larger angles are more frequently associated with positive and smaller angles with negative predictions (see  Fig.~\ref{fig:inter}).

We further attempt to interpret the effect of scaling vector embeddings (mimicking \textit{change in magnitude of flow}). In this context, we hypothesize that small $|s_{i}|$ may indicate that GAV is uncertain whether to flip vector embeddings, which is a necessity for sink/source flow. Therefore, $|s_{i}|$ could be interpreted as a measure of certainty (see Fig.~\ref{fig:inter}, last column). This is confirmed by statistical analysis of $|s_{i}|$, demonstrating that high values ($\mu_{|s_{i}|}$ = 0.62) are assigned to more ($\hat{y}^t_{ij}$ $>$ 0.9 or $<$ 0.1) and low values ($\mu_{|s_{i}|}$ = 0.13) to less certain predictions ($0.4$ $<$ $\hat{y}^t_{ij}$ $<$ $0.6$).

\section{Outlook and Conclusion}
In this work, we propose the simple yet effective Graph Attentive Vectors (GAV) link prediction framework. GAV relies on the idea of modeling simplified physical flow by updating vector embeddings in a constrained manner, which intuitively models the underlying physical process and, therefore, presents a strong inductive bias for link prediction in flow-driven spatial networks. This allows GAV to outperform the previous state-of-the-art by an impressive margin on all metrics across multiple whole-brain vessel and road network datasets while requiring a significantly smaller amount of trainable parameters.
GAV's imitation of the dynamics of physical flow, however, represents a simplified concept, which is not entirely representative of physical principles from, \eg, fluid dynamics (see Fig.~\ref{fig:inter}). Future work should, therefore, aim to further investigate GAV's parameters and extend its assumptions by incorporating different physical principles, such as conservation of mass and momentum, resulting in vector embeddings highly representative of physical flow in flow-driven spatial networks.

{\small
\bibliographystyle{ieee_fullname}
\bibliography{egbib}
}

\onecolumn
\appendix

\renewcommand\thefigure{\arabic{figure}}
\setcounter{figure}{5}
\setcounter{table}{5}

\section{Notations}\label{sup:notation}
We provide a lookup table for notations used in our work.

\begin{table*}[h]
\centering
\caption{Description of notations used in our work.}
\vspace{-1em}
\label{tab:notations}
\renewcommand{\arraystretch}{1.2}
\begin{tabular}{c l} 
\toprule
Notation & Description\\
\midrule
$\mathcal{G}$ & undirected, flow-driven spatial network (or graph)\\
$\mathcal{V}$ & set of nodes in $\mathcal{G}$\\
$n_{i}$ & node $i$ in $\mathcal{G}$\\
$\mathcal{E}$ & set of edges in $\mathcal{G}$\\
$e_{ij}$ & edge (or link) in $\mathcal{G}$ between $n_i$ and $n_j$\\
$\overline{e_{ij}}$ & average edge length estimated over edges in $\mathcal{G}$\\
$\sigma$ & standard deviation of edge length estimated over edges in $\mathcal{G}$\\

$t$ & link prediction target\\
$e^{t}_{ij}$ & target link (or edge) between $n_{i}^{t}$ and $n_{j}^{t}$\\
$n_{i}^{t}$ & target node $i$ affiliated to $e^{t}_{ij}$\\

$h$ & number of hops\\
$\mathcal{G}^{t}_{h}$ & $h$-hop subgraph extracted around $e^{t}_{ij}$\\

$\mathcal{L}(\mathcal{G}^{t}_{h})$ & line graph representation of $\mathcal{G}^{t}_{h}$\\
$\mathcal{V'}$ & set of nodes in $\mathcal{L}(\mathcal{G}^{t}_{h})$\\
$n'_{i}$ & node (or vector embedding) $i$ in $\mathcal{L}(\mathcal{G}^{t}_{h})$\\
$\mathcal{E'}$ & set of edges in $\mathcal{L}(\mathcal{G}^{t}_{h})$\\
$e'_{ij}$ & edge (or link) in $\mathcal{L}(\mathcal{G}^{t}_{h})$ between $n'_{i}$ and $n'_{j}$\\

$k$ & number of message-passing iterations\\
$s_{i}$ & scalar value generated in GAV layer\\
$|s_{i}|$ & absolute value of $s_{i}$ \\
$\tilde{n}'_{i}$ & intermediate node representation (or vector embedding) in GAV layer\\
$\hat{n}'_{i}$ & updated, refined node representation (or vector embedding)\\

$Q$ & query sequence in multi-head attention operation\\
$K$ & key sequence in multi-head attention operation\\
$V$ & value sequence in multi-head attention operation\\

$\hat{y}^t_{ij}$ & GAV's predicted probability of existence of $e^{t}_{ij}$\\
$y^t_{ij}$ & ground truth label of existence of $e^{t}_{ij}$\\

$\mathcal{E}_{\mathcal{N}(n^t_{i})}$ & set of refined vector embeddings originally created from edges adjacent to $n^t_{i}$\\
$N_{i}$ & matrix consisting of $n'_i$ and its direct neighbors\\
$\mathcal{N}(n_{i})$ & set of nodes in the direct neighborhood of $n_{i}$\\
$\mathcal{N}(n_{i}) \cup n_i $ & set of nodes in the direct neighborhood of $n_{i}$ including $n_{i}$ itself\\
$\phi^{(1)}_{\theta}$, $\phi^{(2)}_{\theta}$ & learnable functions in GAV layer\\
$\phi^{(3)}_{\theta}$ & learnable function in readout module\\

$d_{\text{spatial}}$ & spatial dimension (2 or 3)\\
$d_{\text{message}}$ & dimension of $\tilde{n}'_{i}$ in GAV layer\\

$\delta$ & maximum distance threshold utilized in spatial sampling during preprocessing\\
$\mathbin\Vert$ & concatenation operation\\
$\mathcal{L}_{\text{BCE}}$ & binary cross-entropy loss function\\
\bottomrule
\end{tabular}
\end{table*}
\newpage

\section{More on Interpretability and the Modification of Vector Embeddings}\label{sup:inter}
\begin{figure*}[h]
\centerline{\includegraphics[width=\linewidth]{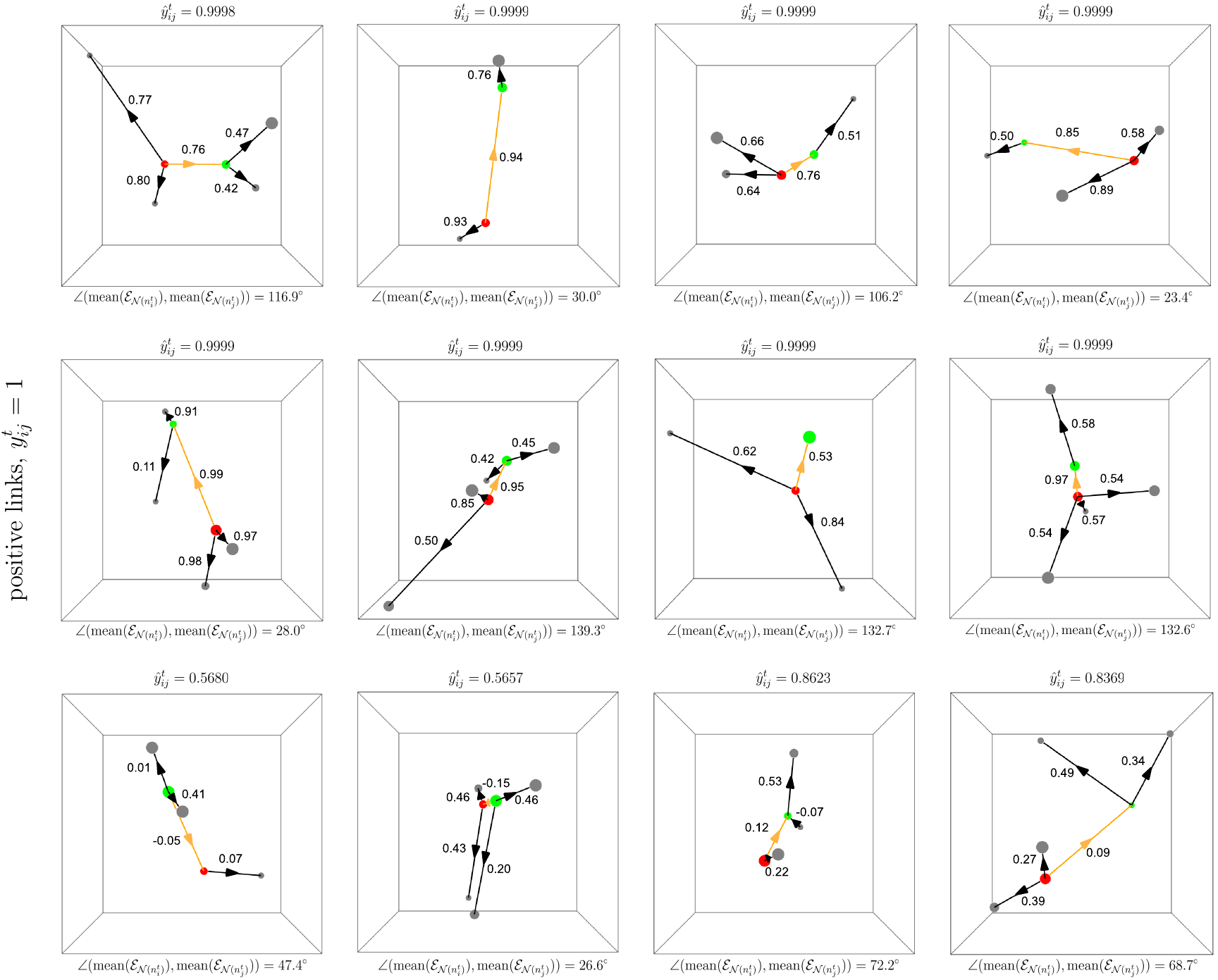}}
\caption{
Visualization of the effect of our GAV layer on vector embeddings. We visualize subgraph representations $\mathcal{G}^t_h$ of 12 positive target links ($y_{ij}^t = 1$) together with the GAV layer's predicted scalar values $s_{i} \in (-1, 1)$. The scalar values $s_{i}$ used to update vector embeddings in $\mathcal{L}(\mathcal{G}^t_h)$ have been projected to their corresponding edges in $\mathcal{G}^t_h$ (see Fig.~\ref{fig:method_overview}) to provide an interpretable visualization. The directionality of edges (indicated by arrows) already incorporates potential shifts in the direction of vector embeddings enforced by our GAV layer. We additionally report the angle $\angle$ between the vector embeddings aggregated around the two target nodes (see Section~\ref{ref:readout}) and the predicted probability of link existence $\hat{y}^t_{ij}$.
}
\label{fig:pos_preds}
\end{figure*}

\begin{figure*}[h]
\centerline{\includegraphics[width=\linewidth]{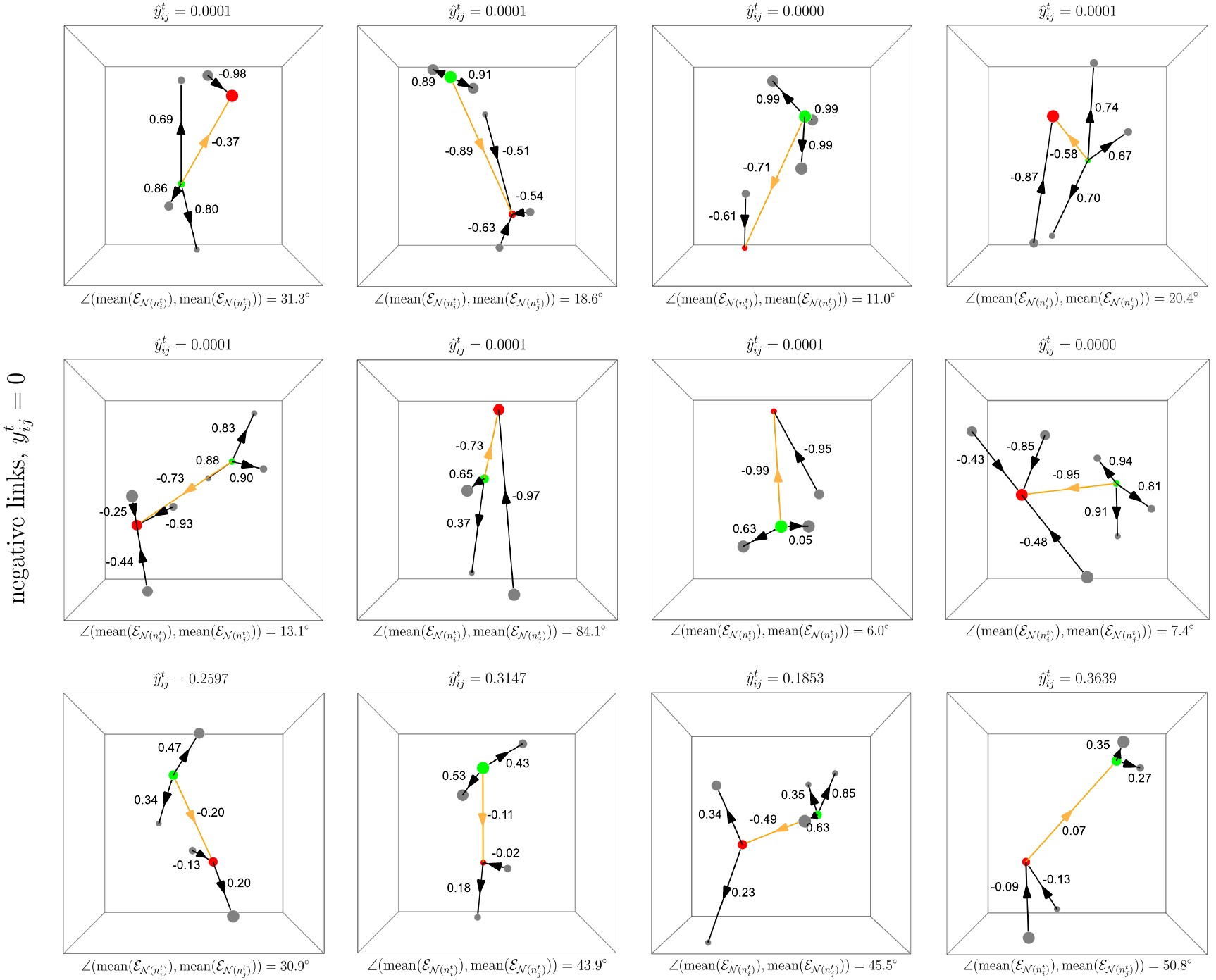}}
\caption{
Visualization of the effect of our GAV layer on vector embeddings. We visualize subgraph representations $\mathcal{G}^t_h$ of 12 negative target links ($y_{ij}^t = 0$) together with the GAV layer's predicted scalar values $s_{i} \in (-1, 1)$. The scalar values $s_{i}$ used to update vector embeddings in $\mathcal{L}(\mathcal{G}^t_h)$ have been projected to their corresponding edges in $\mathcal{G}^t_h$ (see Fig.~\ref{fig:method_overview}) to provide an interpretable visualization. The directionality of edges (indicated by arrows) already incorporates potential shifts in the direction of vector embeddings enforced by our GAV layer. We additionally report the angle $\angle$ between the vector embeddings aggregated around the two target nodes (see Section~\ref{ref:readout}) and the predicted probability of link existence $\hat{y}^t_{ij}$.
}
\label{fig:neg_preds}
\end{figure*}

We provide the interested reader with more visualizations regarding the GAV layer's modification of vector embeddings (see Section~\ref{ref:gav_layer}) on the validation set of the ogbl-vessel benchmark, similar to Fig.~\ref{fig:inter}. These visualizations can also be interpreted as qualitative results. Fig.~\ref{fig:pos_preds} depicts subgraph representations $\mathcal{G}^t_h$ ($h$ set to one) of 12 positive (real, plausible) target links, while Fig.~\ref{fig:neg_preds} depicts subgraph representations of 12 negative (sampled, implausible) target links. Please note that the respective last rows depict challenging cases, as indicated by GAV's predicted probabilities $\hat{y}^t_{ij}$. Additionally, we would like to highlight our hypothesis from Section~\ref{ref:inter} that GAV may attempt to assign the two target nodes (red and green) to sink and source nodes for negative, implausible vessel formations (see Fig.~\ref{fig:neg_preds}), which results in superior representations for link prediction in flow-driven spatial networks that can be effortlessly classified in our physically plausible readout module. Please note that we conduct an additional experiment modifying a toy example in Sec.~\ref{sup:toy} to further facilitate interpretability.\\ \\

\newpage

\section{Initialization of Vector Embeddings}\label{sup:imp_det}
The initialization of the direction of vector embeddings represents an important implementation detail and is based on a straightforward intuition. To be precise, we initialize vector embeddings to point away from the target link $e^{t}_{ij}$, \ie, towards nodes with a node degree of one (leaf nodes). The vector embedding representative of the target link $e^{t}_{ij}$ is set to point from $n_{i}^{t}$ to $n_{j}^{t}$. An exemplary initialization of vector embeddings for a 1-hop subgraph can be found in Fig.~\ref{fig:method_overview}.

\section{GAV and Structural Properties}\label{sup:toy} 
\begin{figure*}[h]
\centerline{\includegraphics[width=0.7\linewidth]{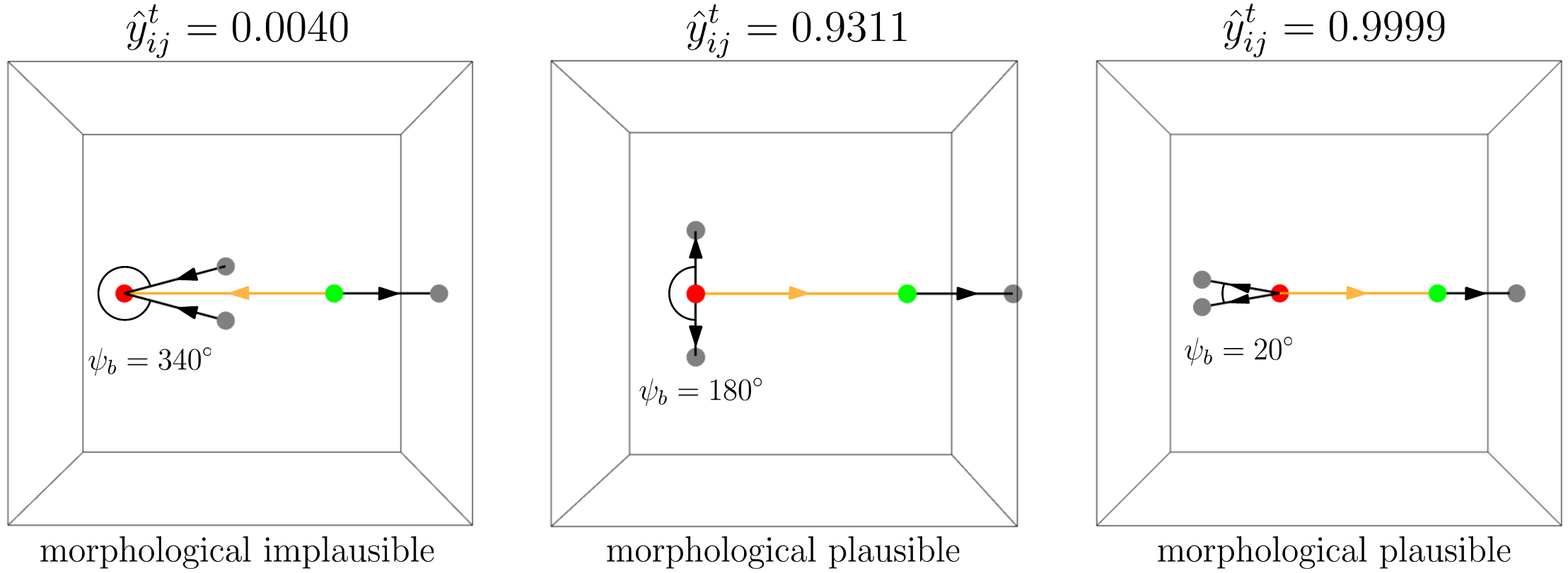}}
\caption{
Morphological implausible (left) and plausible (middle and right) blood vessel formations formed around the target link (orange) with varying bifurcation angles $\psi_b$. GAV correctly identifies morphological plausible blood vessel formations that fulfill relevant hemodynamic functional properties~\cite{schneider2012tissue}.
 }
\label{fig:morp}
\end{figure*}

This section elaborates on how GAV's predictions rely heavily on structural properties, such as bifurcation angles, which reflect functional properties of the underlying physical system~\cite{schneider2012tissue}.
To this end, we prepare and modify a synthetic mock example in Fig.~\ref{fig:morp}. Specifically, we vary the bifurcation angle $\psi_b$ spanned between two edges connected to $n^t_i$ (red) to generate morphological implausible and plausible blood vessel formations (see Fig.~\ref{fig:morp}).

\noindent
As expected, GAV differentiates between plausible and implausible blood vessel formations formed around the target link. GAV assigns a high probability of target link existence $\hat{y}^t_{ij}$ to plausible and a low probability of target link existence to implausible formations. Please note that Fig.~\ref{fig:morp} additionally maps the potentially modified directionality of vector embeddings onto their corresponding edges, similar to Fig.~\ref{fig:pos_preds} and Fig.~\ref{fig:neg_preds}. One can observe the shift in the directionality of vector embeddings created from edges adjacent to nodes with high bifurcation angles $\psi_b$, transforming the target nodes (red and green) to sink and source nodes for morphological implausible blood vessel formations (see Fig.~\ref{fig:morp}, left).

\section{Visualization of Datasets}\label{sup:data}
\begin{figure*}[h]
\centerline{\includegraphics[width=\linewidth]{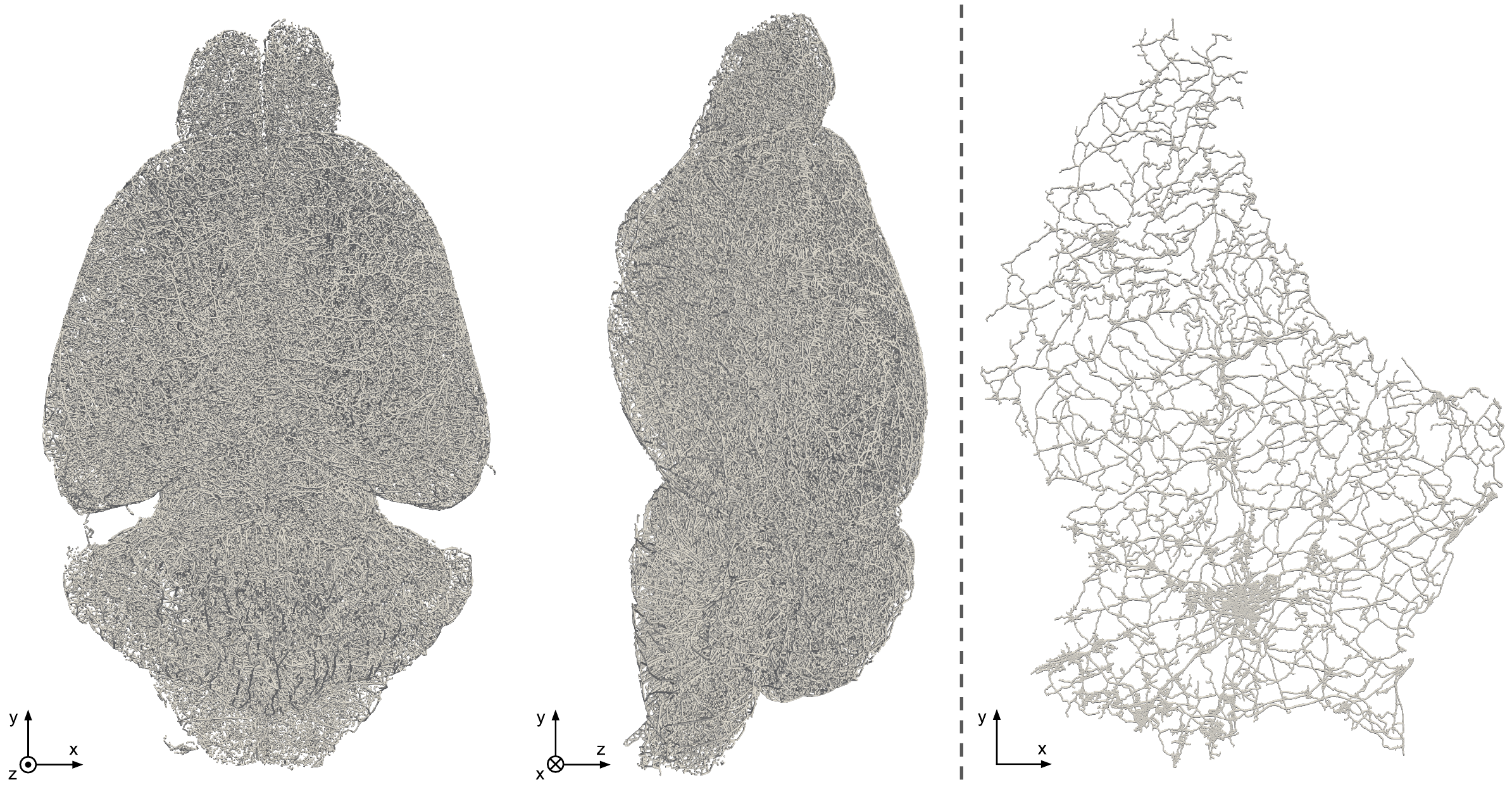}}
\caption{
Visualization of a whole-brain vessel graph and a road network. The depicted flow-driven spatial networks correspond to the raw cd1-tc-vessel and luxembourg-road datasets.
}
\label{fig:data}
\vspace{-1em}
\end{figure*}

In Fig.~\ref{fig:data}, we graphically visualize two flow-driven spatial networks representative of murine whole-brain vessel graphs and road networks.

\section{More Ablations on the GAV Layer}\label{sup:gav_abl}
Since the GAV layer relies on a set of specific design choices, we conduct additional ablation studies determining their influence on the link prediction performance on the validation set of the ogbl-vessel benchmark. To this end, we experiment with different versions of the GAV layer. First, we deactivate the multi-head attention operation; second, we exclude the residual connection; and third, we exchange the leaky ReLU non-linearity in the GAV layer with the ReLU non-linearity. We report our findings in Table~\ref{tab:ablations_des_layer}.

\begin{table}[h]
\centering
\scriptsize
\caption{Ablations on the GAV layer's main design choices.}
\vspace{-1em}
\label{tab:ablations_des_layer}
\begin{tabular}{c c c|c c c c} 
\toprule
Attention & Residual Connection & Leaky ReLU & $\text{AUC}\uparrow$ &$\text{Hits@100}\uparrow$& $\text{Hits@50}\uparrow$& $\text{Hits@20}\uparrow$ \\
\midrule
\cmark & \cmark & \cmark & \cellcolor{teal!40}98.39 & \cellcolor{teal!20}34.46 & \cellcolor{teal!10}26.30 & \cellcolor{teal!40}19.81\\
\xmark & \cmark & \cmark & \cellcolor{teal!10}97.48 & 15.56 & 9.18 & 5.68\\
\cmark & \xmark & \cmark & \cellcolor{teal!20}98.34 & \cellcolor{teal!40}34.90 & \cellcolor{teal!40}27.61 & \cellcolor{teal!20}19.28\\
\cmark & \cmark & \xmark & \cellcolor{teal!20}98.34 & \cellcolor{teal!10}34.16 & \cellcolor{teal!20}26.47 & \cellcolor{teal!10}17.31\\
\bottomrule
\end{tabular}
\end{table}

\noindent
Deactivating the multi-head attention operation (second row) results in a drastic AUC decrease of 0.91, indicating the importance of neighborhood awareness when modifying vector embeddings via our proposed GAV layer. Excluding the GAV layer's residual connection (third row) and using ReLU instead of Leaky ReLU non-linearities (fourth row) leads to a slight reduction in AUC of 0.05, respectively. Based on our reported standard deviation value of $\pm$ 0.02 (see Table~\ref{tab:quantitative_results}), we argue that this performance decrease is indeed significant.

\section{Evaluation Metrics}\label{sup:metrics}
To compare GAV to existing baseline algorithms, we report quantitative results based on the area under the receiver operating characteristic curve (AUC), following the obgl-vessel benchmark. The AUC metric indicates the performance of a classifier by plotting the true positive rate against the false positive rate at all possible classification thresholds. Therefore, AUC provides an aggregate performance measure indicating the classifier's ability to distinguish between positive and negative links.

We introduce the evaluation metric Hits@$k$ as an additional, stricter performance measure. Hits@$k$ compares the classifier's prediction of every single positive link to a randomly sampled set of 100,000 negative links, resulting in a ranking among 100,001 links with respect to the probability of link existence. Based on this ranking, Hits@$k$ indicates the ratio of positive links ranked at the $k$-th place and above. In the context of this work, we evaluate Hits@$k$ at $k=100$, $k=50$, and $k=20$, inspired by other Open Graph Benchmark~\cite{hu2020ogb} link prediction benchmarks.

\section{Configuration of Our Secondary Baseline}\label{sup:sec_base}
We incorporate the EdgeConv message-passing layer~\cite{wang2019dynamic} into the SEAL framework~\cite{zhang2018link, zhang2021labeling}, which has been shown to deliver results on par with or superior to the state-of-the-art on multiple link prediction benchmarks, to introduce a strong, \textit{secondary baseline} (SEAL+EdgeConv) for link prediction in spatial networks. To be precise, we incorporate EdgeConv in SEAL's DGCNN~\cite{zhang2018end}. EdgeConv's update function can be observed in Table~\ref{tab:update}. Here, $\phi_{\theta}$ represents a two-layer MLP with an input dimension of 64, a hidden dimension of 32, and an output dimension of 32. Our modified DGCNN employs in total three EdgeConv layers, with the only difference being that the input dimension of the first EdgeConv layer's MLP corresponds to 70 and the output dimension of the third EdgeConv layer's MLP to 1. Our EdgeConv version utilizes a mean feature aggregation scheme. We set the number of in- and output channels of the DGCNN readout operation's two 1D convolutions to 1 \& 16 and 16 \& 32, respectively. The kernel sizes and strides of the two 1D convolutions correspond to 65 \& 65 and 5 \& 1. We set the input, hidden, and output dimensions of the DGCNN readout operation's MLP to 38, 128, and 1. The global sort pooling layer's parameter k is set to 10. All hyperparameters were tuned on the validation set of the ogbl-vessel benchmark.

\section{GAV's Performance on Non-Flow-Based Link Prediction Benchmarks}\label{sup:non_flow}
To additionally confirm that GAV is specialized for link prediction in flow-driven spatial networks, we conduct an experiment on the ogbl-collab benchmark~\cite{hu2020ogb}, which represents a collaboration network given by an undirected graph where nodes are associated with authors while edges indicate collaborations between them. Node features are comprised of 128-dimensional vectors representative of an author's scientific work (averaged word embeddings reflecting the content of scientific papers). Based on the collaboration network, the task is to predict future collaborations between authors. To adjust GAV to the task of the ogbl-collab benchmark, we model vector embeddings representative of edges in the collaboration network as the difference between 128-dimensional feature vectors of two nodes incident to an edge. We report our findings in Table~\ref{tab:quant_res_collab}.

\begin{table}[h]
\centering
\scriptsize
\caption{Comparison between GAV and SEAL on ogbl-collab and ogbl-vessel. Please note that the increase in GAV's trainable parameters in the experiment on ogbl-collab is mostly due to the increased number of node features (128 vs. 3).}
\vspace{-0.5em}
\label{tab:quant_res_collab}
\begin{tabular}{c|c|l|c} 
\toprule
Dataset & Model & $\#\ \text{Params}\downarrow$ & $\text{Eval. Metric}\uparrow$  (\%)\\
\midrule
\multirow{2}{*}{ogbl-collab}
& SEAL~\cite{zhang2021labeling} & \cellcolor{teal!20}501,570 & \cellcolor{teal!40}64.72 $\text{Hits@50}$\\
& GAV (ours) & \cellcolor{teal!40}44,194 & \cellcolor{teal!20}16.72 $\text{Hits@50}$\\
\midrule
\multirow{2}{*}{ogbl-vessel}
& SEAL~\cite{zhang2021labeling} & \cellcolor{teal!20}172,610 & \cellcolor{teal!20}80.50 $\text{AUC}$\\
& GAV (ours) & \cellcolor{teal!40}8,194 & \cellcolor{teal!40}98.38 $\text{AUC}$\\
\bottomrule
\end{tabular}
\end{table}

\noindent
As expected, GAV, relying on the idea of modeling simplified physical flow in flow-driven spatial networks, does not deliver competitive results on the ogbl-collab benchmark. This is because GAV's strong inductive biases are tailored to link prediction in flow-driven spatial networks and are, therefore, too restrictive for non-flow-based networks, such as ogbl-collab. This repeatedly demonstrates GAV's ability to intuitively model the underlying physical process in flow-driven spatial networks.

To adapt GAV to non-flow-based networks more appropriately, we encourage future work to explore the use of pseudo-spatial positions embedded in nodes of non-flow-based networks rather than its actual node features for the sake of creating vector embeddings. Pseudo-spatial position could, \eg, be determined based on the Fruchterman-Reingold force-directed algorithm~\cite{kobourov2012spring}.

\section{On Translation and Rotation Invariance}\label{sup:invariance}
\begin{wrapfigure}{r}{0.35\textwidth}
  \vspace{-3em}
  \begin{center}
    \includegraphics[width=0.35\textwidth]{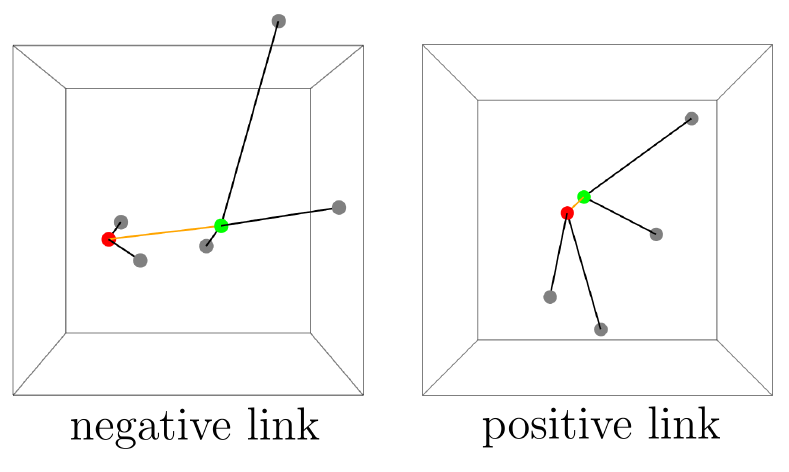}
  \end{center}
  \vspace{-1.5em}
  \caption{Exemplary $\mathcal{G}^{t}_{h}$ extracted from the ogbl-vessel benchmark around a negative and positive link used for experiments in Table~\ref{tab:rot_inv}.}
  \vspace{-1em}
  \label{fig:rot_inv}
\end{wrapfigure}
In this section, we briefly investigate GAV's behavior under rotations and translations of the $h$-hop enclosing subgraph $\mathcal{G}^{t}_{h}$. Specifically, we aim to investigate whether rotation and translation of $\mathcal{G}^{t}_{h}$ result in similar predictions. Since GAV encodes edges as vector embeddings spanned between two nodes (see Section~\ref{ref:subnetw_extraction}), translation invariance is \textit{explicitly ensured}.
However, even though rotation preserves the length and relative angles of edges, rotation invariance is \textit{not explicitly ensured}. This is, \eg, because queries and keys forwarded to the GAV layer's attention operation are not explicitly rotation equi- or invariant, which is one of the key requirements for rotation invariant attention weights and hence potential rotation invariant predictions~\cite{fuchs2020se}. An empirical experiment rotating exemplary input graphs around all three axes, however, demonstrates that GAV's predictions are relatively robust to rotation, indicating to some degree implicit, learned rotation invariance (see Table~\ref{tab:rot_inv}). We encourage future work to further explore the necessity of explicitly encoded translation and rotation invariance in the context of link prediction for flow-driven spatial networks.

\begin{table}[h]
\centering
\scriptsize
\caption{Experiment on rotation invariance of predictions. We rotate three exemplary subgraphs $\mathcal{G}^{t}_{h}$ around all three axes with a step size of 1$^{\circ}$ to investigate GAV's behavior under rotation. We report the standard deviation of predicted target link probability $\hat{y}^t_{ij}$ over all 360 predictions.}
\vspace{-0.5em}
\label{tab:rot_inv}
\begin{tabular}{l|c|c|c} 
\toprule
Subgraph $\mathcal{G}^{t}_{h}$ & $\sigma_{\hat{y}^t_{ij}} \text{(x-axis)}$ & $\sigma_{\hat{y}^t_{ij}} \text{(y-axis)}$ & $\sigma_{\hat{y}^t_{ij}} \text{(z-axis)}$\\
\midrule
Fig.~\ref{fig:morp}, right & 5.15$\cdot e^{-\text{5}}$ & 1.86$\cdot e^{-\text{6}}$ & 1.34$\cdot e^{-\text{5}}$\\
Fig.~\ref{fig:rot_inv}, left & 9.63$\cdot e^{-\text{2}}$ & 1.12$\cdot e^{-\text{3}}$ & 7.64$\cdot e^{-\text{3}}$\\
Fig.~\ref{fig:rot_inv}, right & 1.85$\cdot e^{-\text{3}}$ & 1.08$\cdot e^{-\text{3}}$ & 3.37$\cdot e^{-\text{4}}$\\
\bottomrule
\end{tabular}
\end{table}

\section{Commonalities and Differences between SEAL and GAV}\label{sup:commonalities}
Since the influential SEAL link prediction framework~\cite{zhang2018link, zhang2021labeling} represents one of the most prominent works on learned, GNN-based link prediction algorithms, we would like to clearly state the commonalities and differences between SEAL and our proposed link prediction algorithm tailored to flow-driven spatial networks, GAV. Although GAV utilizes some concepts introduced by SEAL (subgraph extraction/classification \& labeling trick), which are provably used in most competitive approaches and can, therefore, be seen as common practices, we, for the first time, introduce the principle of physical flow to link prediction. To this end, we propose not only a novel flow-inspired, parameter-efficient message-passing layer updating vector embeddings but also a physically plausible readout module facilitating interpretability. Our contributions result in an increase of more than 22\% in AUC compared to SEAL on ogbl-vessel.

\section{Message-Passing Update Functions}\label{sup:update}
We provide a concise overview of message-passing layers featured in our work and their respective high-level, final node update functions in Table~\ref{tab:update}. Here, $d_i$ stands for the node degree of $n_i$, $\alpha_{ij}$ for the learned attention coefficient between $n_i$ and $n_j$, and $\phi_{\theta}$ for an arbitrary learnable function. We would like to highlight the simplicity of the GAV layer's final update function.

\begin{table}[h]
\centering
\caption{Message-passing update functions.}
\vspace{-0.5em}
\label{tab:update}
\renewcommand{\arraystretch}{2}
\begin{tabular}{c | c} 
\toprule
Message-Passing Layer & Update Function\\
\midrule
GAV layer & $\displaystyle \hat{n}_{i} = s_i \cdot n_i$\\

EdgeConv~\cite{wang2019dynamic} & $\displaystyle \hat{n}_{i} = \frac{1}{|\mathcal{N}(n_i)|} \sum_{n_j \in \mathcal{N}(n_i)} \phi_{\theta}(n_{i} \mathbin\Vert n_{j} - n_{i})$\\
GAT layer~\cite{velivckovic2017graph} & $\displaystyle \hat{n}_{i} = \alpha_{ii} \cdot \phi_{\theta}(n_{i}) + \sum_{n_j \in \mathcal{N}(n_i)} \alpha_{ij} \cdot \phi_{\theta}(n_{j})$\\

SAGE layer~\cite{hamilton2017inductive} & $\displaystyle \hat{n}_{i} =  \phi_{\theta}^{(1)}(n_{i}) +  \phi_{\theta}^{(2)}(\frac{1}{|\mathcal{N}(n_i)|} \sum_{n_j \in \mathcal{N}(n_i)} n_{j})$\\

GCN layer~\cite{kipf2016semi} & $\displaystyle \hat{n}_{i} =  \phi_{\theta}(  \sum_{n_j \in \mathcal{N}(n_i)\, \cup\, n_i} \frac{1}{\sqrt{d_i \cdot d_j}}\, n_j)$\\
\bottomrule
\end{tabular}
\end{table}

\section{Medical Relevance of the Link Prediction Task for Whole-Brain Vessel Graphs}\label{sup:relevance}
Since GAV has been developed around the ogbl-vessel benchmark, we would like to provide more details on the medical relevance and the application of link prediction algorithms for whole-brain vessel graphs. As already mentioned in Section~\ref{sec:intro}, vascular network representations of the brain originate from a multi-stage, imperfect process, typically consisting of a segmentation stage followed by a graph extraction stage (skeletonization and pruning). Detailed pipelines for whole-brain vessel graph generation can be found in the literature~\cite{paetzold2021whole, walchli2021hierarchical, meyer2009voreen, drees2021scalable}. Each stage of the graph generation pipeline introduces noise and artifacts to the extracted whole-brain vessel graphs. The initial segmentation stage~\cite{todorov2020machine}, \eg, often results in under- or over-connected vessel segmentation masks, which in turn result in equally under- or over-connected whole-brain vessel graphs. This is mostly due to the shortage of annotated training data (especially in the 3D domain), which is required for accurate vessel segmentation via supervised state-of-the-art deep-learning-based segmentation techniques.
Under-/over-connectivity, however, limits the application of whole-brain vessel graphs for subsequent medically relevant downstream tasks, such as the diagnosis, treatment, and analysis of neurovascular brain disorders (\eg, aneurysms or strokes). This is because these downstream tasks require flawlessly connected whole-brain vessel graphs free of artifacts to obtain a deeper understanding of neurovascular brain disorders by, \eg, accurately recognizing anomalies in blood flow patterns via blood flow modeling~\cite{schmid2021severity}.
To overcome the obstacle of under-/over-connectivity in whole-brain vessel graphs and, therefore, to enable researchers to obtain a more accurate and advanced understanding of neurovascular brain disorder, one can either optimize whole-brain vessel graph generation pipelines~\cite{shit2022relationformer, shit2021cldice} or utilize the task of link prediction, which we extensively investigate in this work.

\end{document}